\documentclass[runningheads]{llncs}

\usepackage{eccv}

\usepackage{eccvabbrv}

\usepackage{graphicx}
\usepackage{booktabs}

\usepackage[accsupp]{axessibility}  

\usepackage{graphicx}
\usepackage{amsmath}
\usepackage{amssymb}
\usepackage{booktabs}
\usepackage{balance}
\usepackage{lipsum}
\usepackage{caption}
\usepackage{algorithm}
\usepackage{algorithmic}
\usepackage{enumerate}
\usepackage{engord}
\usepackage{bm}
\usepackage{bbm}
\usepackage{amstext}
\usepackage{comment}
\usepackage{color}
\usepackage{booktabs}
\usepackage{multirow}
\usepackage{mathtools}
\usepackage{setspace}
\usepackage{pifont}
\usepackage{dsfont}
\usepackage{cancel}
\usepackage[inline]{enumitem}
\usepackage[toc,page]{appendix}
\usepackage{hhline}
\usepackage{adjustbox}
\usepackage{makecell}

\usepackage{hyperref}

\usepackage[capitalize]{cleveref}
\definecolor{Gray}{gray}{0.9}
\definecolor{White}{gray}{1}
\definecolor{WhiteGray}{rgb}{0.9, 0.9, 0.9}
\definecolor{DGray}{gray}{0.8}
\definecolor{DDDDGray}{gray}{0.3}
\definecolor{citecolor}{HTML}{0071bc}

\definecolor{DeltaColor}{rgb}{0.039,0.73,0.71}
\definecolor{SigmaColor}{rgb}{0.98,0.45,0.0}
\definecolor{AlphaColor}{rgb}{0,0,0.8}
\definecolor{BetaColor}{rgb}{0.8,0,0.8}
\definecolor{GammaColor}{rgb}{0.514,0.34,0.224}
\definecolor{EpsilonColor}{rgb}{0.353,0.725,0.906}
\definecolor{GreenColor}{rgb}{0.137,0.573,0.565}
\definecolor{RedColor}{rgb}{0.949,0.275, 0.224}
\definecolor{Cardinal}{rgb}{0.549,0.082,0.082}

\DeclareMathAlphabet\mathbfcal{OMS}{cmsy}{b}{n}

\newcommand{\colorRef}[1]{\textcolor{black}{#1}}
\crefname{figure}{\colorRef{Fig.}}{\colorRef{Figs.}}
\Crefname{figure}{\colorRef{Figure}}{\colorRef{Figures}}
\crefname{section}{\colorRef{Sec.}}{\colorRef{Secs.}}
\Crefname{section}{\colorRef{Section}}{\colorRef{Sections}}

\crefname{table}{\colorRef{Tab.}}{\colorRef{Tabs.}}
\Crefname{table}{\colorRef{Table}}{\colorRef{Tables}}
\Crefname{equation}{\colorRef{Eq.}}{\colorRef{Eqs.}}
\Crefname{equation}{\colorRef{Equation}}{\colorRef{Equation}}

\newcommand\method{GOMP\xspace}
\newcommand{\qheading}[1]{\noindent\mbox{\textbf{#1}\;}}

\usepackage{orcidlink}

\begin{document}

\title{Grasp-Oriented Non-Prehensile Manipulation via Learning a Graspability Field} 

\titlerunning{GOMP}

\author{Licheng Zhong\orcidlink{0009-0003-3645-7870} \and
Gim Hee Lee\orcidlink{0000-0002-1583-0475}}

\authorrunning{L. Zhong and G. H. Lee}

\institute{Department of Computer Science, National University of Singapore, Singapore \\
\email{\{zlicheng, gimhee.lee\}@comp.nus.edu.sg} \\
\url{https://zlicheng.com/gomp_page/}}

\maketitle

\begin{abstract}
  Non-prehensile manipulation is often used as a preparatory step for robotic grasping, yet existing approaches typically require a predefined target object pose. 
In practice, however, objects admit multiple graspable configurations and the desired pose is not known in advance. 
We reformulate non-prehensile manipulation for grasping as optimizing an object centric graspability objective rather than reaching a specific pose. 
We construct a graspable set from synthesized grasps and define a graspability field that measures how suitable an object configuration is for successful grasp execution. 
The scalar measure provides a dense learning signal for reinforcement learning and determines when to terminate manipulation. This yields a closed-loop manipulation-to-grasp pipeline driven by a single policy.
Experiments in simulation and on a real robot show that the policy reliably reconfigures objects into graspable states and transitions to grasping without external planners or manually specified stopping conditions. 
The predicted graspability distance correlates with real world grasp success, 
which indicates that the learned representation captures grasp feasibility of object configurations.
\end{abstract}

\begin{figure}[!h]
  \centering
  \includegraphics[width=\linewidth]{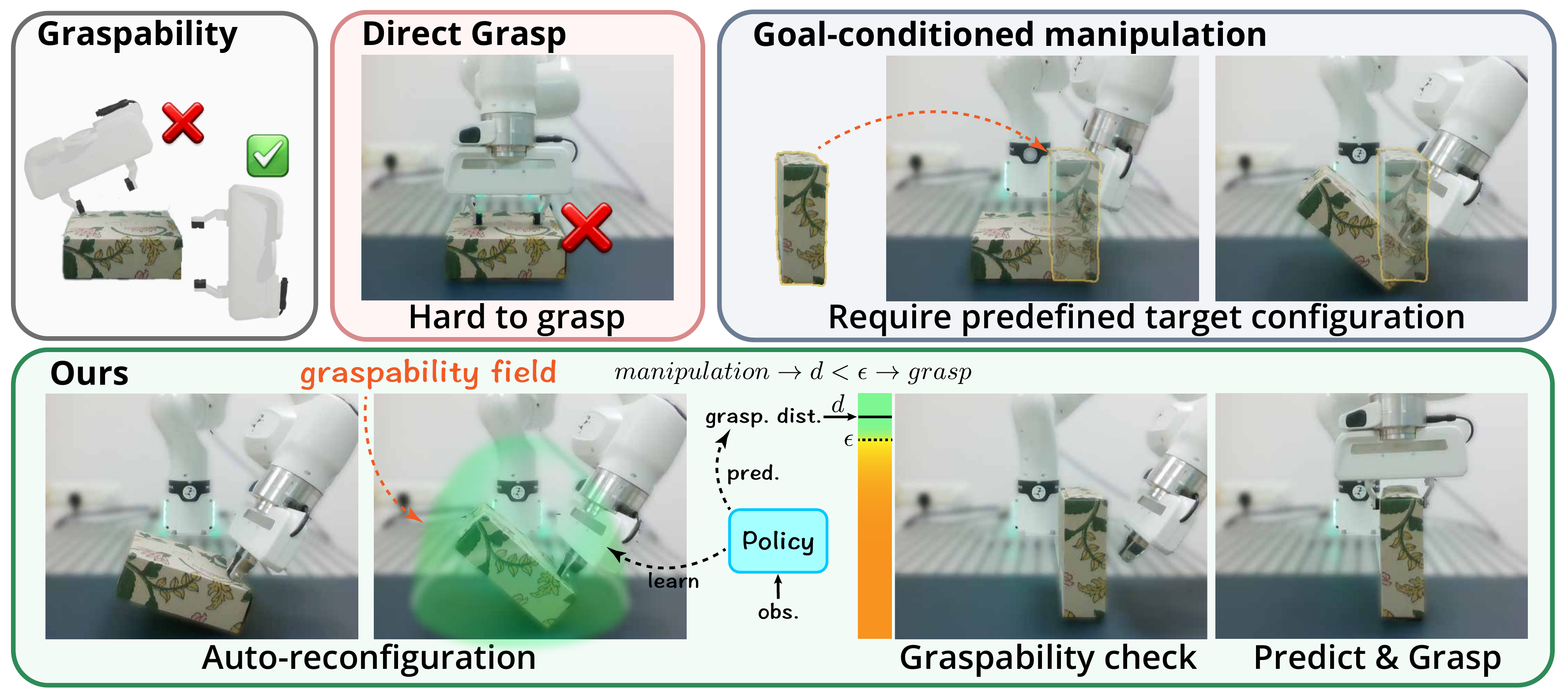}
  \caption{Direct grasps from a fixed approach can fail due to unfavorable object configurations, even when feasible grasps exist.
Instead of requiring a predefined target pose, we learn a graspability field over object states to guide manipulation and trigger the manipulation-to-grasp transition. Once the object enters a graspable region, the robot autonomously transitions from manipulation to grasp execution.}
  \label{fig:teaser}
\end{figure}

\section{Introduction}
\label{sec:intro}

Robotic grasping in unstructured environments often fails not because feasible grasps do not exist, but because the object is initially in an unfavorable configuration.
Before a reliable grasp can be executed, the robot frequently needs to interact with the object through pushing, sliding, or reorienting it.
These non-prehensile interactions serve as a preparation stage that transforms the object into a configuration suitable for grasping, especially when the object is unstable, partially occluded, or poorly aligned with the gripper.

Recent learning-based approaches have shown promising results by using non-prehensile manipulation as a preparation stage for grasping.
Methods such as CORN~\cite{cho2024corn} and DyWA~\cite{lyu2025dywa} formulate manipulation as a goal-conditioned control problem, where the policy is guided toward a desired object pose.
This formulation enables stable learning and has demonstrated effective object repositioning behaviors in both simulation and real-world settings.

However, the target-pose formulation implicitly assumes that grasp preparation can be reduced to reaching a single predefined configuration.
In practical grasping scenarios, many objects admit multiple graspable configurations, and selecting one goal configuration does not necessarily reflect how a robot should interact with the object from its current state.
A policy may successfully match the target pose while failing to meaningfully improve grasp success.

More importantly, even when a target pose is specified, reaching it does not necessarily yield the most graspable configuration: a pose-reaching objective optimizes geometric similarity to a predefined configuration, while grasp preparation requires moving toward a nearby region of high graspability.
Since graspability typically exhibits multiple local maxima in the object configuration space, a predefined target may lie outside the nearest graspable basin, forcing the policy to move the object away from an easily graspable configuration before eventually approaching the target. As a result, manipulation behavior ends up guided by distance to the target rather than improvement of grasp success.

This reveals a mismatch between pose-reaching objectives and grasp preparation: preparing an object for grasping is not a pose regulation task, but a state improvement process, where the policy should be guided by how graspable the object currently is rather than its proximity to a predefined pose.
As shown in~\cref{fig:teaser}, we therefore argue that non-prehensile manipulation for grasping should be driven by graspability rather than explicit target poses: manipulation is a process of progressively improving the graspability of the object, with the goal not of reaching a particular configuration, but of transforming the object into a state that admits reliable and stable grasps.

Based on this formulation, we propose \textbf{\method}, a \textbf{G}rasp-\textbf{O}riented non-prehensile \textbf{M}anipulation \textbf{P}olicy that maximizes a learned graspability function defined over object configurations.
Instead of predicting or tracking a target pose, we learn a graspability field that estimates how suitable the current configuration is for executing a grasp.
The policy is trained using reinforcement learning to progressively increase this graspability over time, which naturally enables both manipulation behaviors and stopping decisions for grasp execution.

Experiments in simulation and on a real Franka robot demonstrate that \method consistently improves grasp success over target-conditioned baselines.
In summary, our contributions are threefold:
\begin{itemize}
\item We introduce a new formulation of non-prehensile manipulation as graspability optimization, 
which replaces pose-reaching objectives with a state-based improvement objective.

\item We propose a graspability field representation that provides dense learning signals and enables both manipulation and stopping without requiring explicit target poses.

\item We validate the proposed approach through extensive simulation studies on non-prehensile manipulation, and real-world closed-loop experiments that demonstrate consistent improvements in grasp success when using our grasp-oriented manipulation policy as a preparation stage.

\end{itemize}

\section{Related Work}
\label{sec:related}

\qheading{Non-Prehensile Manipulation.}
\label{sec:related_nonprehensile}
Non-prehensile manipulation studies how a robot deliberately changes the configuration of an object without grasping it. Typical actions include pushing, sliding, and reorientation~\cite{mason1999progress,lynch1996stable,wang2025dexterous,lee2024non,zhong2025activepusher,jung2025spin}.
Both classical planning approaches and contact-based control methods aim to transform an object from an initial state to a desired configuration~\cite{cheng2022contact,mordatch2012discovery,mordatch2012contact}.
Recent learning-based methods acquire such behaviors directly from interaction data. 
Self-supervised and reinforcement learning approaches learn pushing and rearrangement policies~\cite{zeng2018learning}, including push–grasp coordination strategies that improve robustness~\cite{dogar2010push,xu2021efficient,hu2025push}. 
Despite improved adaptability, the manipulation objective in these methods remains externally specified, typically through a target pose or goal conditioned policy.
More recent work investigates coordinated whole arm reorientation~\cite{zhou2023hacman,jiang2024hacmanpp,cho2024corn,lyu2025dywa,zhu2025adaptpnp}. 
CORN~\cite{cho2024corn} learns policies that manipulate objects toward a predefined target pose, and DyWA~\cite{lyu2025dywa} further improves perception and distillation. 
To achieve strong performance, these methods require the desired object configuration to be known in advance.
However, the desired configuration is usually unknown in practical grasping scenarios.
Objects often admit multiple feasible grasp poses, and perception alone cannot determine which configuration should be reached before interaction. 
Therefore, for grasping, the objective is not to reach a specific pose but to bring the object into any configuration from which a stable grasp can be executed.

\smallskip
\qheading{Learning Based Robotic Grasping.}
\label{sec:related_grasping}
Learning-based robotic grasping aims to predict grasp actions directly from sensory observations~\cite{huang2025robograsp,deng2025graspvla}. 
Early methods learn grasp success predictors or analytic grasp synthesis models from large-scale data, often assuming precise object geometry or full visibility~\cite{wang2023dexgraspnet,zhang2024dexgraspnet}. 
Vision-based approaches then train networks to estimate stable grasp poses or grasp affordances from images or point clouds~\cite{guo2016deep,sundermeyer2021contact,fang2023anygrasp,fang2019graspnet,fang2020graspnet,Wang_2021_ICCV}. 
Recent generative models produce diverse grasp candidates conditioned on object observations, 
which improve robustness to shape variation and perception noise~\cite{urain2022se3dif,weng2023neural,hager2021graspme,jiang2021synergies}.
Although effective for grasp execution, these approaches generally assume that the object is already in a graspable configuration when grasp poses are predicted. 
Some work considers interaction with the environment to make specific grasp poses reachable, such as extrinsic dexterity where objects are pushed against surfaces to enable occluded grasps~\cite{zhou2023learning}. 
These methods still aim to execute a given grasp by modifying reachability. 
In contrast, our objective is to prepare the object itself by bringing it into any configuration that admits stable grasps without specifying a target pose.

\smallskip
\qheading{Affordance Guided Manipulation.}
\label{sec:related_affordance}
Several works attempt to connect manipulation and grasping through grasp-related predictions~\cite{pohl2020affordance,hundt2020good}.
Vision-based grasping systems often estimate grasp affordances, reachability, or success likelihoods from sensory observations~\cite{ding2024preafford,yuan2025robopoint,li2025learning,sundermeyer2021contact,breyer2021volumetric,Wang_2021_ICCV,mahler2017dex}. 
Manipulation actions are then chosen to expose or enable predicted grasps~\cite{zeng2018learning,xu2024naturalvlm,li2024learning,lin2025affordance}.
These approaches still require a grasp to be selected or evaluated during planning. 
The objective is to realize a particular grasp candidate or maximize predicted grasp success, rather than to optimize the object configuration itself.
In contrast, our method does not predict or evaluate grasps during manipulation. 
We instead define a graspability measure over object states and learn actions that transform the object toward configurations that admit stable grasps. 
Thus, manipulation is guided by improving the object state rather than enabling a specific grasp.

\section{Problem Formulation}
\label{sec:problem_formulation}
\begin{figure}[t]
  \centering
  \includegraphics[width=\linewidth]{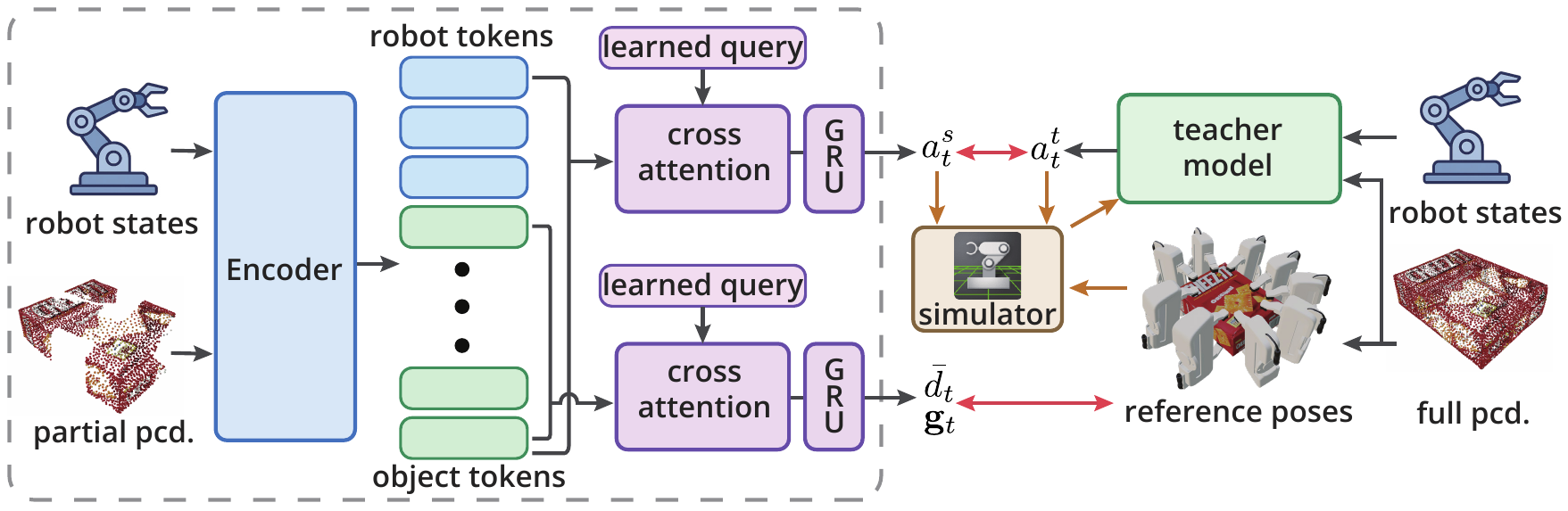}
  \caption{Method achitecture. A teacher–student framework that encodes observations, aggregates temporal information with a recurrent module, and predicts manipulation actions and a graspability distance. The teacher is used only during training, and the robot executes a grasp once the predicted graspability indicates a feasible configuration.
  }
  \label{fig:pipeline}
\end{figure}

\noindent \textbf{Goal.} We study non-prehensile manipulation as a preparatory process for robotic grasping, where the goal is to transform an object into a configuration that admits reliable grasps. 

\smallskip
\noindent \textbf{Formulation.} Let $s_o^t$ denote the (unobserved) rigid 6D pose $(R_o, p_o)$ of the object at time $t$, 
and the policy only receives an observation $o_t$ derived from sensor measurements such as point clouds.  We consider an observation-based manipulation policy $\pi(a_t \mid o_t)$ that outputs non-prehensile actions $a_t$.
Instead of specifying a desired target pose, we consider that an object may admit multiple graspable configurations. 
We therefore define a \emph{graspable set} $\mathcal{S}_g$, which contains object states that allow stable grasps. 
Manipulation is formulated as moving the object toward this set 
instead of toward a single predefined pose.

To quantify progress toward graspability, we define a scalar graspability measure $G(s_o)$ that evaluates how close a state is to the graspable set $\mathcal{S}_g$. 
Given an initial state $s_o^0$, the manipulation policy aims to produce a trajectory $\{s_o^t\}_{t=0}^T$ that maximizes the graspability of the final state:
\begin{equation}
    \pi^* = \arg\max_\pi \mathbb{E}_\pi \left[ G(s_o^T) \right],
    \label{eq:objective}
\end{equation}
where the expectation is taken over the state distribution induced by the policy and the environment dynamics.
This formulation eliminates the need for an explicit target pose: any state within the graspable set is considered successful. 
In the following section, we show how the graspable set can be constructed and how $G(s_o)$ is derived from distances to this set.

\section{Our Method}
\label{sec:method}

\noindent \textbf{Overview.} \cref{fig:pipeline} shows the overview of our framework. 
Reference graspable configurations from full object geometry (\cref{sec:refer_pose}) are used only during training to 
build graspability field for supervision (\cref{sec:grasp_field}) and reward (\cref{sec:rewards}), and are not available at test time.
From partial observations, the policy predicts manipulation actions, a graspability distance, and a grasp pose (\cref{sec:gomp}).
The robot manipulates in closed loop and executes a grasp once the predicted distance indicates a sufficiently graspable configuration (\cref{sec:grasp_transition}), removing the need for a predefined target pose.

\smallskip
\noindent \textbf{Architecture.}
As shown in~\cref{fig:pipeline}, we employ a perception–control architecture 
that estimates graspability and actions jointly. 
Observations are encoded using the object-centric encoder from CORN~\cite{cho2024corn}. 
Since our tasks and objects differ from the original setting, we fine-tune only the final attention layer while freezing earlier layers 
to preserve geometric features while adapting to new interaction dynamics.
The encoder produces object and robot tokens, which are queried through cross-attention conditioned on the robot state to obtain a compact representation of the current object configuration.

Due to partial observability of the object configuration from a single frame, the student policy incorporates a recurrent module (GRU~\cite{cho2014gru}) to aggregate temporal information.
The recurrent state allows the policy to infer object motion from recent observations and actions, 
which improves robustness to occlusion and contact-induced dynamics. 
The resulting feature is passed to the prediction heads that output the manipulation action, the predicted graspability distance, and the terminal grasp pose.
A teacher policy with a similar perception backbone but without recurrent memory is used only to generate stable manipulation behaviors for distillation.

\subsection{Graspable Reference Pose Generation}
\label{sec:refer_pose}

The graspable set $\mathcal{S}_g$ is not directly available in practice and cannot be specified analytically for arbitrary objects. 
To approximate this set, we employ an off-the-shelf grasp synthesis model~\cite{urain2022se3dif} that predicts feasible grasps from object geometry.

Given an observed object point cloud, the grasp model produces multiple candidate grasps.
Each grasp corresponds to a relative pose between the gripper and the object.
We convert each grasp into a reference object pose by fixing a vertical top-down gripper pose and inverting the grasp transform $T^{\mathrm{ref}}_{\mathrm{obj}} = R_x(\pi)\,\bigl(T^{\mathrm{obj}}_{\mathrm{grasp}}\bigr)^{-1}$,
where $R_x(\pi)$ is a $180^\circ$ rotation about the $x$-axis for vertical top-down alignment.
The object is then vertically translated to rest on the table.
This results in a set of graspable object states:
\begin{equation}
    \mathcal{S}_g=\{\hat{s}_o^i\}_{i=1}^G,
    \label{eq:graspable_set}
\end{equation}
where each $\hat{s}_o^i$ represents a configuration in which a stable grasp can be executed.

Importantly, these configurations are not used as target poses for the manipulation policy. 
They are never provided to the policy as observations, nor used for imitation or supervision. 
Instead, they serve solely to define a set of desirable object states that characterize graspability.

In the next section, we derive a scalar graspability measure by computing distances between the current object state and this graspable set.

\subsection{Graspability Field Definition}
\label{sec:grasp_field}

A graspable configuration is not unique: many distinct object poses may allow stable grasps. 
Therefore, measuring progress toward grasping using a single target pose is inappropriate 
since it artificially constrains the manipulation objective. 
Instead, we measure how close the current object configuration is to a set of graspable states.

Given the graspable set $\mathcal{S}_g=\{\hat{s}_o^i\}_{i=1}^{G}$ defined in \cref{sec:problem_formulation}, we define a distance from the current state $s_o$ to this set.
A natural measure of graspability is the minimum distance to the set:
\begin{equation}
d_{\mathrm{anchor}}(s_o)=\min_{i\in\{1,\ldots,G\}}\mathcal{D}(s_o,\hat{s}_o^i),
\label{eq:dist_anchor}
\end{equation}
where $\mathcal{D}(\cdot,\cdot)$ measures the geometric discrepancy between two object configurations. 
In practice, $\mathcal{D}$ compares the spatial arrangement of the object geometry (\eg, keypoints or surface points) under two configurations, and does not depend on the robot end-effector pose. 
A smaller value indicates that the object is closer to a configuration that admits a stable grasp.

In our implementation, we approximate $\mathcal{D}$ using geometry alignment. 
Let $\{p_k\}_{k=1}^{K}$ denote a fixed set of keypoints sampled from the object bounding box. 
For a configuration $s_o$ with rigid transform $T(s_o)$, the distance between two configurations is computed as the mean Euclidean discrepancy of transformed keypoints:
\begin{equation}
\mathcal{D}(s_o,\hat{s}_o^i)
=
\frac{1}{K}\sum_{k=1}^{K}
\|T(s_o)p_k - T(\hat{s}_o^i)p_k\|_2.
\end{equation}
This formulation compares object geometry rather than pose parameters and does not require an explicit pose representation at inference time. 
In particular, the policy only observes point clouds, while the geometric distance is used internally to construct the learning signal.

However, directly optimizing the minimum distance leads to discontinuities 
since the closest reference configuration may switch abruptly during manipulation. 
This instability makes learning difficult. 
To obtain a smooth and differentiable objective, we replace the hard minimum with a soft aggregation:
\begin{equation}
d_{\mathrm{soft}}(s_o)=-\tau\log\sum_{i=1}^G\exp\left(-\frac{\mathcal{D}(s_o,\hat{s}_o^i)}{\tau}\right),
\label{eq:dist_soft}
\end{equation}
which smoothly approximates the minimum distance. 
As $\tau\rightarrow0$, $d_{\mathrm{soft}}$ approaches $d_{\mathrm{anchor}}$, while larger $\tau$ allows the measure to account for multiple nearby graspable configurations simultaneously.

Although the soft distance provides a smooth objective, it may still encourage frequent switching between different graspable modes across time steps. 
To promote temporal consistency, we introduce an anchor term based on the previously selected configuration index $i^*$:
\begin{equation}
\tilde{d}(s_o^t)=d_{\mathrm{soft}}(s_o^t)+\lambda\mathcal{D}(s_o^t,\hat{s}_o^{i^*}).
\label{eq:dist_combined}
\end{equation}
The anchor steers the object toward the selected graspable configuration as it approaches the target. This mechanism stabilizes the trajectory and prevents oscillations between competing grasp modes.

We interpret $\tilde{d}(s_o)$ as a graspability field defined over the space of object configurations. 
States with smaller values correspond to higher graspability. 
Importantly, this field evaluates a property of the object state 
rather than alignment with any specific grasp pose. 
Specifically, the policy does not aim for a particular end-effector pose. Instead, it seeks to move the object into a configuration that admits many feasible grasps.
Unlike target-based objectives that optimize distance to a single pose, this field defines a continuous landscape guiding the object toward regions that allow reliable grasps. 
Consequently, multiple manipulation strategies that produce graspable object states are equally valid.

\subsection{Learning with Graspability-Driven Rewards}
\label{sec:rewards}

The graspability field defined in Sec.~\ref{sec:grasp_field} provides a continuous measure of how suitable an object configuration is for grasping. 
To learn a manipulation policy, we convert this measure into a training signal.
A straightforward choice is to define the instantaneous reward as a monotonic function of the graspability distance:
\begin{equation}
r_t=f\left(\tilde{d}(s_o^t)\right),
\label{eq:reward}
\end{equation}
where smaller distance corresponds to higher reward. 
This directly aligns policy learning with the objective of reducing the distance to the graspable set.

However, optimizing a state-based reward can lead to slow learning due to delayed credit assignment. 
To encourage incremental progress toward graspable configurations, we employ potential-based reward shaping. 
Following previous work~\cite{cho2024corn}, we use a potential function as a monotonic decreasing function of the graspability distance:
\begin{equation}
\Phi(s_o) \triangleq \phi(\tilde d(s_o)), \qquad 
\phi(d)=k_1\,\beta^{k_2 d},
\label{eq:potential_def}
\end{equation}
where $k_1$ and $k_2$ control the scale and sharpness of the potential and $\beta\in(0,1)$.
The shaped reward is then defined as:
\begin{equation}
r_t=\gamma\Phi(s_o^{t+1})-\Phi(s_o^t),
\label{eq:potential_reward}
\end{equation}
which rewards reductions in the graspability distance between consecutive states.

Intuitively, the agent receives a positive reward whenever the object moves closer to the graspable set (\ie, when $\tilde d$ decreases). This provides a dense supervision signal and eliminates the need to execute grasp attempts during training.
This formulation encourages the policy to progressively move the object toward graspable configurations while preserving the optimal policy under the original objective. 
Importantly, the reward is not a heuristic signal but a direct consequence of the graspability field: learning corresponds to minimizing the distance to the graspable set over time.

\subsection{Grasp-Oriented Manipulation Policy (\method)}
\label{sec:gomp}

We instantiate the proposed formulation using a grasp-oriented manipulation policy, termed \method. 
The policy maps observations to non-prehensile manipulation actions and is trained to optimize the graspability field defined in \cref{sec:grasp_field}.
Instead of assuming access to the full object state, the policy operates directly on sensory observations. 
At each time step $t$, the policy receives an observation $o_t$ derived from perception, including point cloud measurements and robot proprioceptive information, and produces an action
\begin{equation}
a_t \sim \pi_\theta(\cdot \mid o_t),
\label{eq:policy}
\end{equation}
where $a_t$ corresponds to a non-prehensile manipulation command.
In addition to actions, the policy predicts a scalar graspability distance $\hat{d}_t$ and a grasp pose used only for the final grasp execution. 
Neither reference configurations nor any target pose is provided as input.

We emphasize that the policy architecture itself is not specific to our method. 
Any policy capable of mapping observations to manipulation actions can be used within our framework 
since the key contribution lies in the graspability-based objective 
instead of the network design.
%
During training, the predicted distance $\hat d_t$ is supervised to regress the graspability distance $\tilde d(s_o^t)$ defined by the graspability field.
This enables the policy to estimate graspability directly from observations and to use the same signal for both manipulation and the manipulation-to-grasp transition at execution time.

\subsection{Graspability-Guided Manipulation-to-Grasp Transition}
\label{sec:grasp_transition}

Unlike target-based manipulation, our formulation does not specify a desired object pose. 
Consequently, the system must determine when the object is sufficiently prepared for grasping and when manipulation should terminate.
We leverage the graspability field to provide this decision signal.
Although the graspability field is defined over object states during training, the system relies on the policy-predicted graspability distance $\hat d_t$ at execution time.
Manipulation is considered complete once the predicted graspability indicates that the object has reached a sufficiently graspable configuration.

Since the magnitude of the graspability distance depends on object scale, a fixed threshold would vary across objects of different sizes.
To obtain a scale-invariant stopping criterion, we normalize the predicted distance by the object size.
Let $L$ denote the diagonal length of the object bounding box.
We define the normalized predicted graspability distance as $\bar d_t=\hat d_t / L$.
%
We transition from manipulation to grasp execution when the graspability distance remains below a threshold for a short temporal window:
\begin{equation}
\bar d_{t:t+K} < \epsilon,
\label{eq:stop}
\end{equation}
where $\bar d_{t:t+K}$ denotes the maximum of the predicted normalized graspability distances over a temporal window of $K$ steps.



Importantly, the transition criterion does not rely on a predefined target pose, and thus the policy can decide when to stop manipulating. Any configuration in the graspable set counts as success. Since GOMP optimizes over the entire graspable set rather than a single target, the policy can move toward alternative graspable regions that are more stable and reachable under contact dynamics; geometrically graspable but dynamically unstable states do not satisfy the stability constraint and are therefore not counted as successful. At the transition time, the policy outputs a grasp pose $\mathbf{g}_t$ to parameterize the final grasp action. The pose does not supervise manipulation and does not define the learning objective. It only specifies how the gripper should execute the grasp once the object reaches a graspable configuration.
Using the same graspability measure for both learning (Sec.~\ref{sec:rewards}) and stopping teaches the policy how to manipulate the object and when further manipulation is unnecessary.
This produces a closed-loop pipeline in which manipulation naturally prepares the object for grasp execution.

\section{Experiment}
\label{sec:exp}

\subsection{Experimental Setup \& Implementation Details}
\label{sec:exp_setup}

\noindent \textbf{Simulation Setup.}
All training is done in IsaacSim~\cite{NVIDIA_Isaac_Sim} and IsaacLab~\cite{Mittal_Isaac_Lab}, using a Franka Emika Panda with a parallel jaw gripper.
For both training and evaluation, we manually select task-suitable objects from the DexGraspNet~\cite{wang2023dexgraspnet}, GraspFactory~\cite{srinivas2025graspfactory}, and the YCB dataset~\cite{calli2015benchmarking,calli2017yale,calli2015ycb}.
The teacher policy receives full object point cloud, 
while the student policy operates on partial observations. Following CORN~\cite{cho2024corn}, we render partial point clouds using nvdiffrast~\cite{laine2020modular}.

\begin{figure}[t]
    \centering
    \begin{subfigure}[c]{0.48\linewidth}
        \centering
        \includegraphics[width=0.85\linewidth]{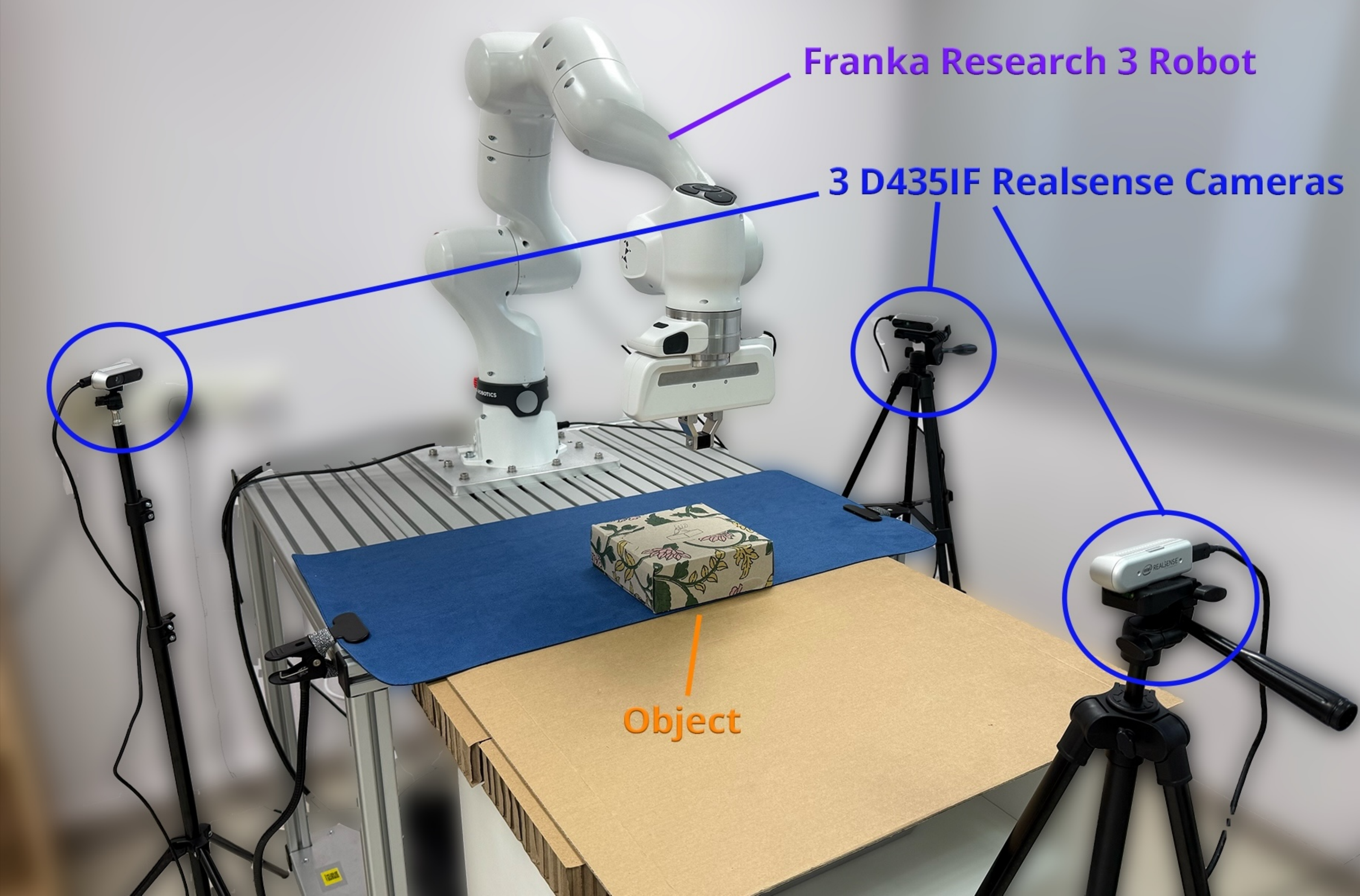}
        \caption{Real world setup}
        \label{fig:real_setup_main}
    \end{subfigure}
    \hfill
    \begin{subfigure}[c]{0.46\linewidth}
        \centering
        \includegraphics[width=0.85\linewidth]{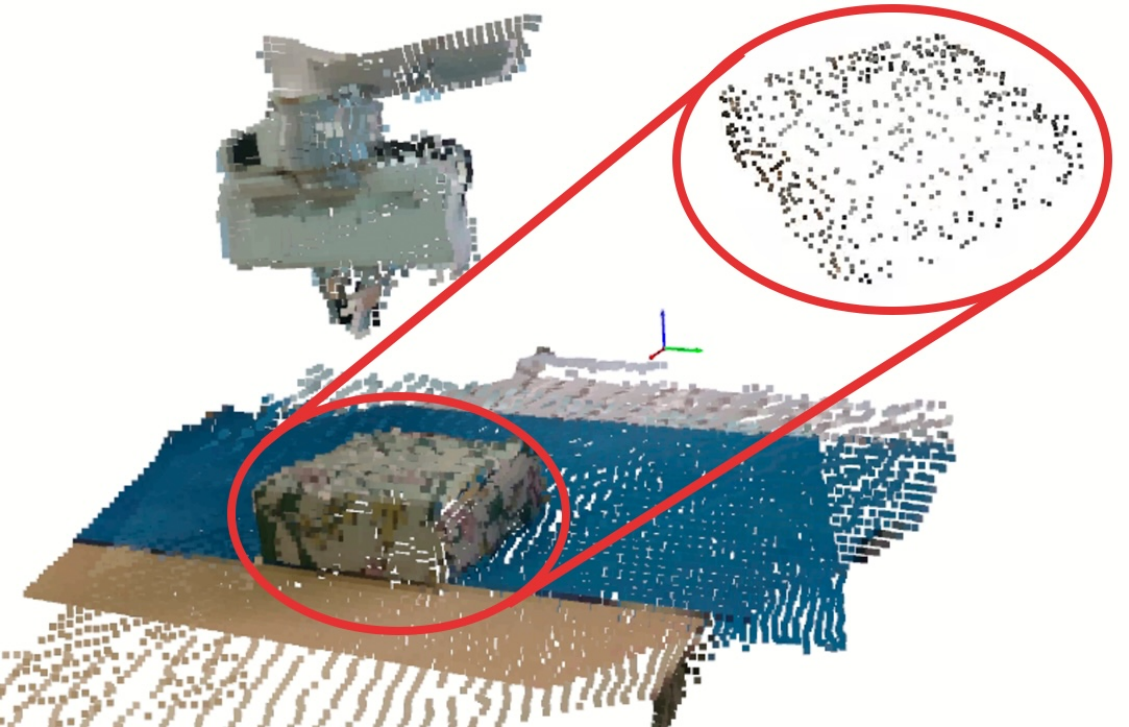}
        \caption{Perception input}
        \label{fig:real_setup_pcd}
    \end{subfigure}

    \caption{
Real-world experimental setup \textbf{(a)} and perception input \textbf{(b)}.
}
    \label{fig:real_setup}
\end{figure}
\begin{figure}[t]
  \centering
    \includegraphics[width=\linewidth]{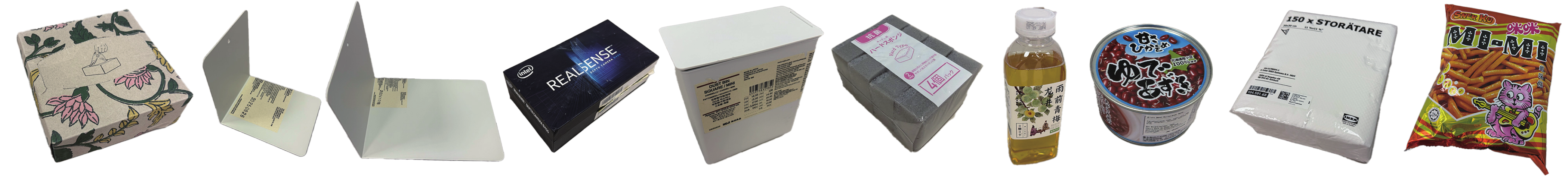}
  \caption{Objects used in real world experiments.
  }
  \label{fig:real_objects}
\end{figure}

\smallskip
\noindent \textbf{Real World Setup.}
\label{sec:exp_real_setup}
As shown in~\cref{fig:real_setup}, we run real-world experiments on a Franka Research 3 with a parallel jaw gripper. Three Intel RealSense D435IF RGB-D cameras capture multi-view depth images that are fused into a single partial point cloud. The cloud is filtered by a workspace bounding box, and the robot points are removed using forward kinematics~\cite{cho2024corn}.
As illustrated in~\cref{fig:real_objects}, the evaluation set includes objects with diverse geometries, friction properties, and compliance levels, and no ground-truth poses, markers, or external tracking systems are used.
All experiments use the standard parallel jaw gripper without friction enhancing materials. Successful grasps therefore depend on the policy 
that brings objects into graspable configurations, not on extra gripper friction.

\smallskip
\noindent \textbf{Implementation Details.}
We train our policy with a teacher-student framework: reinforcement learning followed by online policy distillation.
The teacher policy is trained for 15k iterations (approximately 10 hours on a single NVIDIA RTX A6000 GPU).
A velocity constraint on the object is introduced after 5k iterations.
Before policy distillation, we further train the teacher for an additional 10k iterations with an action-scale curriculum to reduce the effective action range.
Domain randomization is applied throughout training, including variations in object mass, scale, friction, and initial poses.


\subsection{Non-Prehensile Manipulation Evaluation}
\label{sec:exp_nonprehensile}










\begin{table}[t]
\centering
\footnotesize
\setlength{\tabcolsep}{4pt}
\renewcommand{\arraystretch}{1.12}
\caption{
Simulation evaluation of non-prehensile manipulation as a grasp preparation task.
Target Pose indicates whether the policy is explicitly conditioned on a predefined object configuration.
Stable Check indicates whether dynamic stability constraints (velocity thresholds) are required in the success criterion.
}
\begin{adjustbox}{max width=\linewidth}
\begin{tabular}{l cc cc cc}
\toprule

\multirow{2}{*}{Method} 
& \multirow{2}{*}{\makecell{Target\\Pose}} 
& \multirow{2}{*}{\makecell{Stable\\Check}} 
& \multicolumn{2}{c}{Seen Objects} 
& \multicolumn{2}{c}{Unseen Objects} \\

\cmidrule(lr){4-5}\cmidrule(lr){6-7}

& 
&
& Success Rate $\uparrow$ 
& Avg. Time $\downarrow$ 
& Success Rate $\uparrow$ 
& Avg. Time $\downarrow$ \\

\midrule

Retrained Baseline~\cite{cho2024corn,lyu2025dywa} & \ding{51} &\ding{55} & 36.1\% & 5.3s & 28.8\% & 6.1s \\
Nearest-reference Oracle & \ding{51} &\ding{55} & 52.0\% & 8.1s & 50.2\% & 8.8s \\
Multi-reference Oracle & \ding{51} &\ding{55} & 42.2\% & 11.3s & 39.4\% & 9.9s \\
Pretrained Baseline~\cite{cho2024corn,lyu2025dywa} & \ding{51}&\ding{55}  & -- & -- & 6.1\% & 12.9s \\
Pretrained Baseline~\cite{cho2024corn,lyu2025dywa} & \ding{51} &\ding{51} & -- & -- & 1.9\% & 12.1s \\
\method (ours) Teacher & \ding{55} &\ding{51} & 84.0\% & 4.6s & 77.9\% & 4.8s \\
\midrule
\method (ours) Student & \ding{55} &\ding{51} & 75.5\% & 6.1s & 68.1\% & 6.2s \\

\bottomrule
\end{tabular}
\end{adjustbox}

\label{tab:sim_exp}
\end{table}

We first evaluate non-prehensile manipulation in simulation, where precise object states are available. This experiment does not assess grasp success; it tests whether the policy can reliably reconfigure objects toward graspable configurations without predefined target poses.

\smallskip
\noindent \textbf{Baselines.}
We compare with the goal conditioned pose reaching teacher model used in both CORN~\cite{cho2024corn} and DyWA~\cite{lyu2025dywa}, whose original evaluation targets physically stable resting poses verified in a different simulator (IsaacGym).
Direct comparison is non-trivial because these methods target pose reaching, not grasp preparation: our reference poses come from grasp synthesis and are graspable but not always stable resting poses.
Thus, we evaluate:
1) \textbf{Pretrained baseline}: the released CORN/DyWA teacher checkpoints on our objects, with targets sampled from our (possibly dynamically unstable) synthesized references, making pose reaching difficult even when feasible grasps exist.
2) \textbf{Retrained baseline}: the same teacher policy reimplemented and retrained in our codebase with identical observations and action space, isolating the effect of the objective from implementation details; the unstable-target issue remains.
3) \textbf{Multi-reference oracle}: the retrained baseline evaluated against all candidate references per episode, reporting the best result, which upper-bounds the gain from simply having access to multiple targets.
4) \textbf{Nearest-reference oracle}: the retrained baseline using, at every step, the reference closest to the current object state as its target -- the fairest upper bound, since it always selects the most reachable target.
5) \textbf{Our method}: our target-free policy that optimizes the graspability objective defined in~\cref{sec:method}. We use $10$ reference configurations in all evaluations.

\smallskip
\noindent \textbf{Evaluation metric.}
We evaluate manipulation as grasp preparation, so success requires both geometric alignment and dynamic stability: position error $<0.05$m, orientation error $<15^\circ$, linear velocity $<0.01$m/s, and angular velocity $<0.1$rad/s.
The goal-conditioned baseline targets its assigned reference pose, while GOMP -- not conditioned on a target pose -- succeeds if the final configuration reaches \emph{any} graspable reference within the same tolerances; both therefore share the same geometric criterion, with GOMP simply having access to the full graspable set rather than a single target.
For completeness, we also report the geometric criterion used by prior work (without velocity thresholds).

\smallskip
\noindent \textbf{Results.}
Tab.~\ref{tab:sim_exp} shows that goal conditioned pose reaching degrades when the targets are unstable (pretrained baseline); this does not indicate a failure of the original method, but reveals a mismatch between pose-reaching objectives and grasp preparation.
Retraining the same policy in our setup improves performance but still remains substantially worse, often oscillating due to residual pose error.
In contrast, our policy achieves high success on seen/unseen objects without any target pose and terminates once the configuration becomes sufficiently graspable.
This experiment only evaluates the manipulation stage: goal conditioned methods require an external grasp planner once the target pose is reached, while our policy also predicts when to transition to grasp execution.
We evaluate the closed-loop pipeline on a real robot in~\cref{sec:exp_real}.

\subsection{Manipulation-to-Grasp Evaluation}
\label{sec:exp_real}

\begin{figure}[t]
  \centering
    \includegraphics[width=1.0\linewidth]{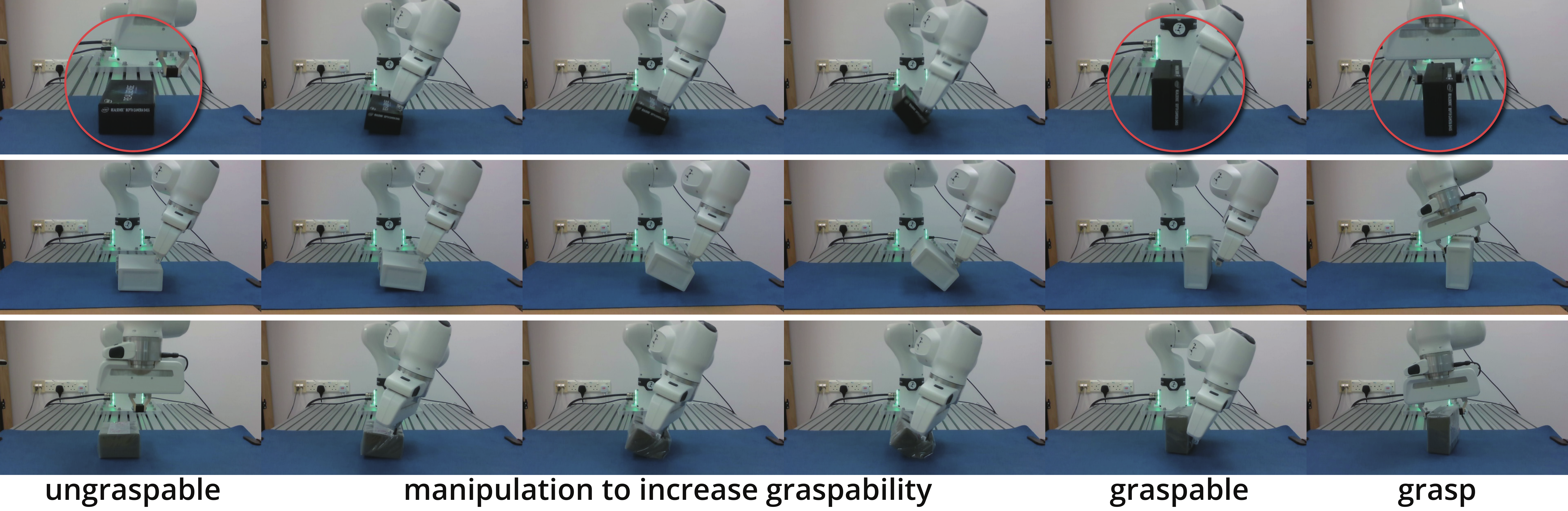}
  \caption{Qualitative results of the full manipulation-to-grasp pipeline on real world objects. Each row shows a temporal sequence where the robot manipulates the object to increase graspability and autonomously transitions to grasp execution once a graspable configuration is reached.
  }
  \label{fig:real_results}
\end{figure}

We evaluate the full manipulation-to-grasp pipeline on a real robot (Sec.~\ref{sec:exp_real_setup}).
For each object, we manually set challenging initial configurations where the object is in a non-graspable state, so that a direct grasp from a fixed approach pose without prior manipulation almost always fails ($8 \%$ success).
The same policy performs non-prehensile manipulation until the predicted graspability distance falls below the transition threshold, and then executes a grasp using its predicted grasp pose (no auxiliary planners or external detectors).
Quantitative and qualitative results are shown in~\cref{tab:real_results,fig:real_results}.

The subsequent grasps are almost always successful once non-prehensile manipulation succeeded. Failures are mainly due to unsuccessful reconfiguration, which indicates that the graspability field provides a reliable signal for deciding when manipulation has sufficiently prepared the object for grasping.
The only consistently unsuccessful object is \emph{Snack}, which is highly deformable and slippery.
For rigid objects such as \emph{PaperBox} and \emph{DustBin}, failures arise from two modes: (1) the object slides during manipulation, causing insufficient or inaccurate reconfiguration within the time limit; and (2) the robot reaches the predefined safety force limit due to strong contact with the table or object, terminating manipulation before a graspable configuration is reached.
The predicted graspability distance still correlates with grasp success when manually placed in a favorable configuration.

\begin{table}[t]
\centering
\footnotesize
\setlength{\tabcolsep}{3.2pt}
\renewcommand{\arraystretch}{1.05}
\caption{Real world manipulation-to-grasp performance on diverse objects.
Each object is tested five times from challenging initial configurations.
}
\begin{adjustbox}{max width=\linewidth}
\begin{tabular}{lccccccc c cc c}
\toprule
& PaperBox & SmallBookend  & BigBookend & RealsenseBox & DustBin
& Sponge & Bottel & Can & TissuePack & Snack &  Avg.\\
\midrule
Manip.
& 4/5 & 4/5 & 4/5 & 5/5 & 3/5 & 4/5 & 4/5 & 3/5
& 1/5 & 0/5 &  64\%(32/50) \\
Grasp
& 4/4 & 4/4 & 3/4 & 5/5 & 3/3 & 4/4 & 4/4 & 3/3
& 1/1 & 0/0 &   97\%(31/32) \\
\midrule
Total
& 4/5 & 4/5 & 3/5 & 5/5 & 3/5 & 4/5 & 4/5 & 3/5
& 1/5 & 0/5 &  62\%(31/50) \\
\bottomrule
\end{tabular}
\end{adjustbox}

\label{tab:real_results}
\end{table}

\begin{figure}[t]
    \centering
    \begin{subfigure}[c]{0.48\linewidth}
        \centering
        \includegraphics[width=\linewidth]{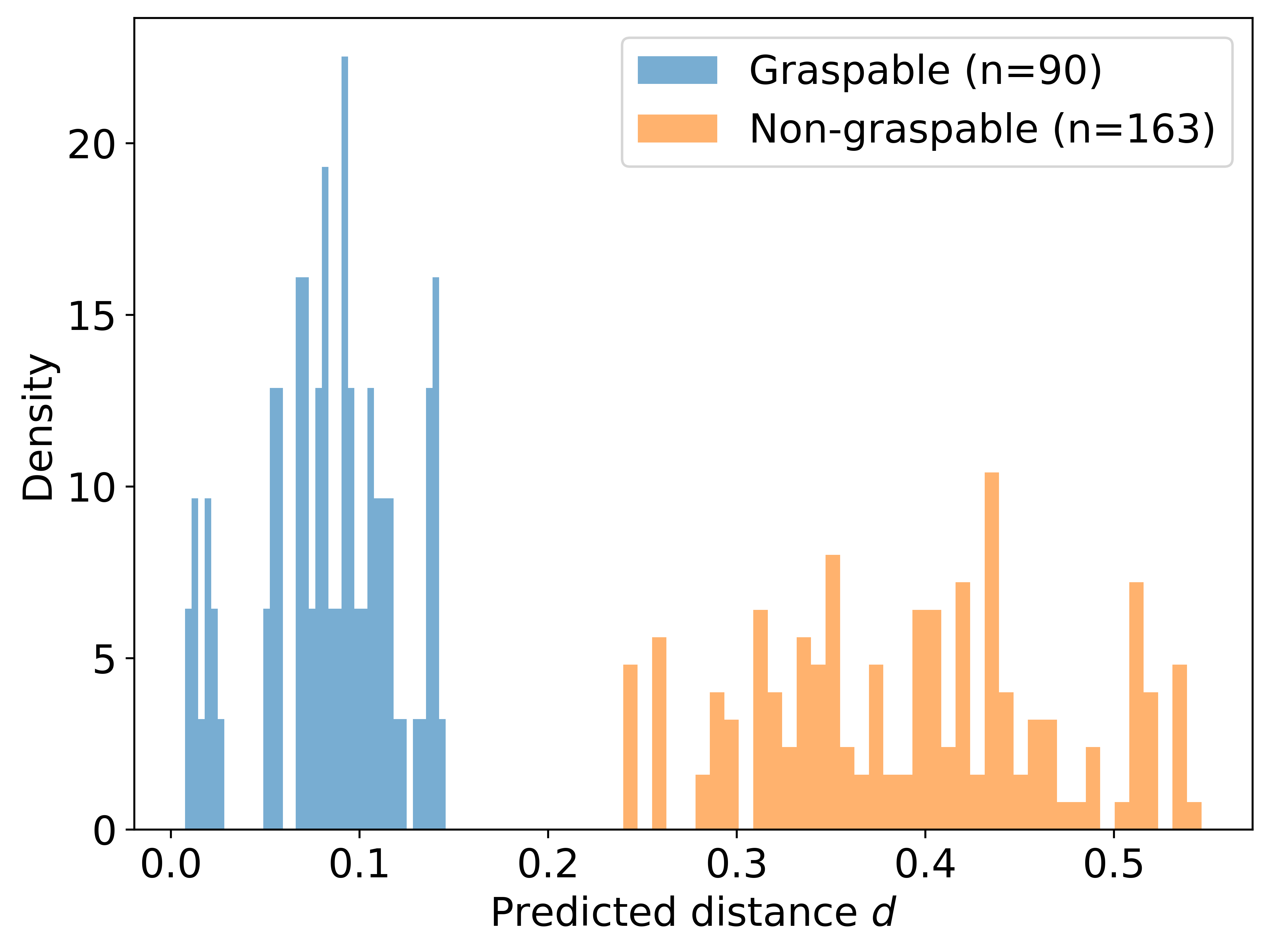}
        \caption{Distribution of predicted graspability distance.}
        \label{fig:grasp_dist}
    \end{subfigure}
    \hfill
    \begin{subfigure}[c]{0.46\linewidth}
        \centering
        \includegraphics[width=\linewidth]{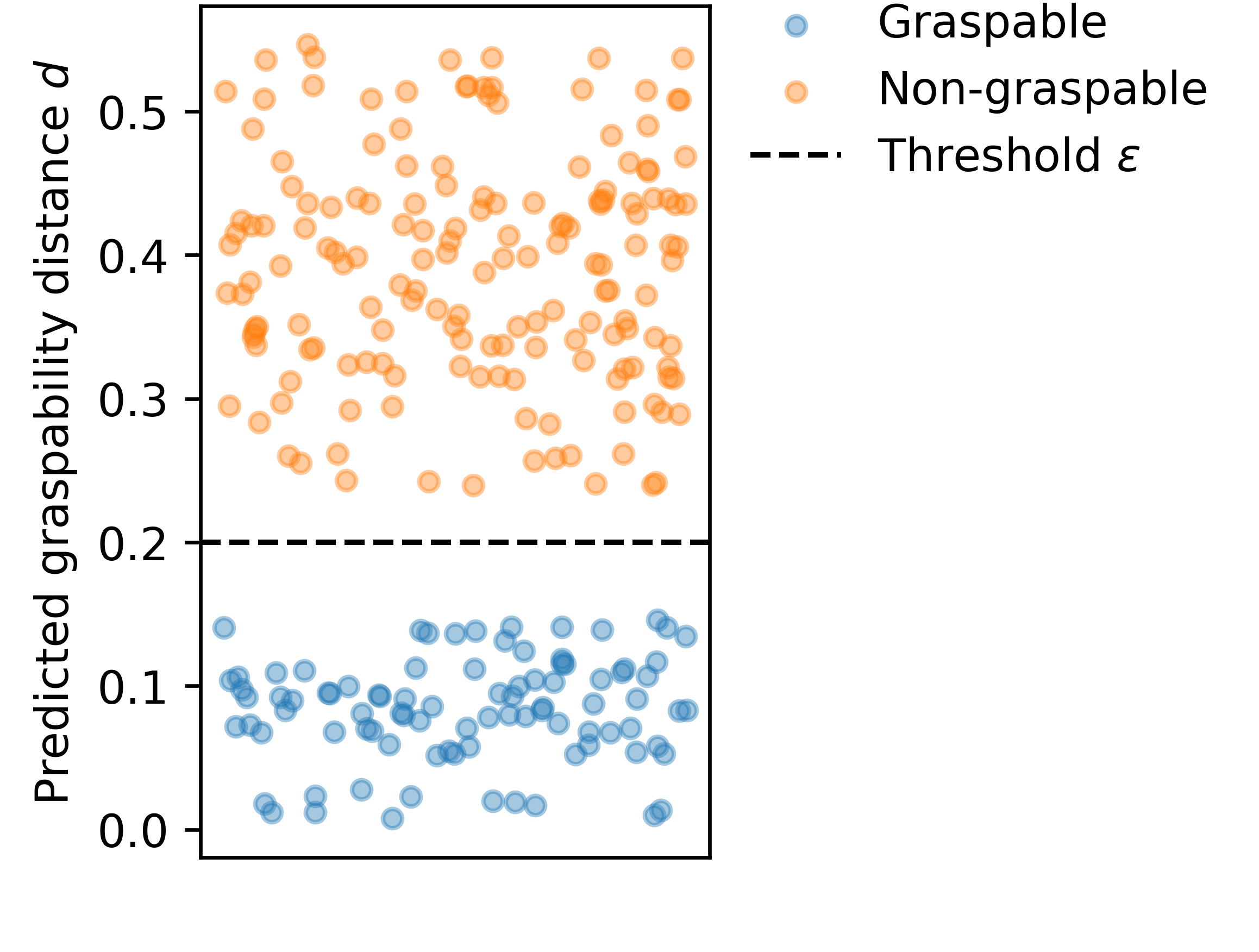}
        \caption{Graspability-based separation and transition threshold.}
        \label{fig:grasp_sep}
    \end{subfigure}

    \caption{
    Validation of graspability-based transition in real-world experiments.
    Graspable configurations correspond to lower predicted distances; a single threshold separates graspable from non-graspable states.
    }
    \label{fig:transition_eval}
\end{figure}

\qheading{Validation of Graspability Based Transition.}
We validate the predicted graspability distance in~\cref{eq:stop} across diverse real world object configurations.
Labeling each configuration as graspable or non-graspable reveals a clear separation in predicted distances (Fig.~\ref{fig:transition_eval}), supporting a single transition threshold $\epsilon$.
Instead of reflecting the executed motion, the result indicates that the predicted distance captures a property of the object configuration. This justifies its use as both a learning objective and the manipulation-to-grasp transition signal.

\subsection{Ablation Studies}
\label{sec:exp_ablation}

We ablate the number of reference poses and the graspability distance design (hard-min vs.\ soft aggregation, and removing the temporal anchoring term in~\cref{eq:dist_combined}).
Tab.~\ref{tab:ablation} shows that increasing the number of references generally improves performance on unseen objects, 
and thus indicates better coverage of graspable configurations. Using $10$ references provides the best trade-off between success rate and execution time, while too many references slightly slow execution and may reduce success due to conflicting descent directions.

Interestingly, even a single reference pose already yields reasonable performance. This does not reduce the objective to target-pose reaching: supervision is defined by distance to a grasp-feasible configuration rather than reproducing a specific pose. 
Due to contact and support constraints, configurations far from the reference pose can still be graspable while some configurations with small SE(3) error are not.
The policy therefore learns to move the object into a graspable basin instead of matching a particular configuration; multiple references mainly improve coverage of this basin, while a single reference still provides a locally valid descent direction.
In contrast, a pose-tracking reward continues to penalize pose error even after the object becomes graspable. This often induces corrective motions or oscillations.

\begin{table}[t]
\centering
\footnotesize
\setlength{\tabcolsep}{3.2pt}  
\renewcommand{\arraystretch}{1.05}
\caption{Ablation studies in simulation.}
\begin{adjustbox}{max width=\linewidth}
\begin{tabular}{lcccccc}
\toprule
& \multicolumn{3}{c}{Seen Objects} & \multicolumn{3}{c}{Unseen Objects} \\
\cmidrule(lr){2-4}\cmidrule(lr){5-7}
 & Success Rate $\uparrow$ & Avg. Time $\downarrow$ & Final $\bar d_T$ $\downarrow$ & success rate $\uparrow$ & Avg. Time $\downarrow$ & Final $\bar d_T$ $\downarrow$ \\

\midrule
1 refer. poses  & 82.2\% & 4.6s & 0.095 & 72.8\% & 4.8s & 0.115 \\
3 refer. poses  & 80.8\% & 4.7s & 0.102 & 75.2\% & 5.1s & 0.107 \\
5 refer. poses  & 76.2\% & 5.6s & 0.112 & 71.4\% & 5.8s & 0.122 \\
8 refer. poses  & 81.2\% & 4.8s & 0.102 & 75.6\% & 5.2s & 0.111 \\
\textbf{10 refer. poses} & \textbf{84.0\%} & \textbf{4.6s} & 0.098 & \textbf{77.9\%}& \textbf{4.8s} & 0.118  \\
15 refer. poses & 79.1\% & 4.8s & 0.106 & 72.1\% & 5.3s & 0.121 \\
20 refer. poses & 76.0\% & 6.4s & 0.115 & 68.1\% & 6.0s & 0.132 \\
\midrule
use hardmin~\cref{eq:dist_anchor} & 81.1\% & 5.0s & 0.102 & 74.3\% & 5.2s & 0.118 \\
w/o anchor~\cref{eq:dist_combined} & 78.1\% & 5.0s & 0.108 & 74.1\% & 5.1s & 0.116 \\

\bottomrule
\end{tabular}
\end{adjustbox}

\label{tab:ablation}
\end{table}

\section{Conclusion}
\label{sec:conclu}

We proposed a grasp-oriented formulation of non-prehensile manipulation for robotic grasping. 
Instead of reaching a predefined target pose, the robot is trained to optimize an object-centric graspability objective. 
We construct a graspable set from synthesized grasps and define a graspability field that measures how suitable an object configuration is for grasp execution. 
This scalar measure provides a unified signal for learning manipulation behaviors and decides when manipulation should terminate, which enables a closed-loop manipulation-to-grasp pipeline with a single policy.
Experiments in simulation and the real world show that the policy reliably reconfigures objects into graspable states and transitions to grasping without external planners or manually specified stopping conditions. 
The same graspability measure is used for learning, transition control, and evaluation, and consistently predicts real-world grasp success. 
This suggests the learned distance represents grasp feasibility of object configurations instead of a task-specific reward heuristic.

\section*{Acknowledgements}
This research / project is supported by the National Research Foundation (NRF) Singapore, under its NRF-Investigatorship Programme (Award ID. NRF-NRFI09-0008), and the Tier 2 grant MOET2EP20124-0015 from the Singapore Ministry of Education.

%
%
\bibliographystyle{configs/splncs04}
\bibliography{refers}

\clearpage
\appendix
\section*{Appendix}
\section{Additional Results}
\subsection{Qualitative Manipulation Simulation Results}
\label{sec:supp_sim_qual}

\begin{figure}[h]
  \centering
    \includegraphics[width=1.0\linewidth]{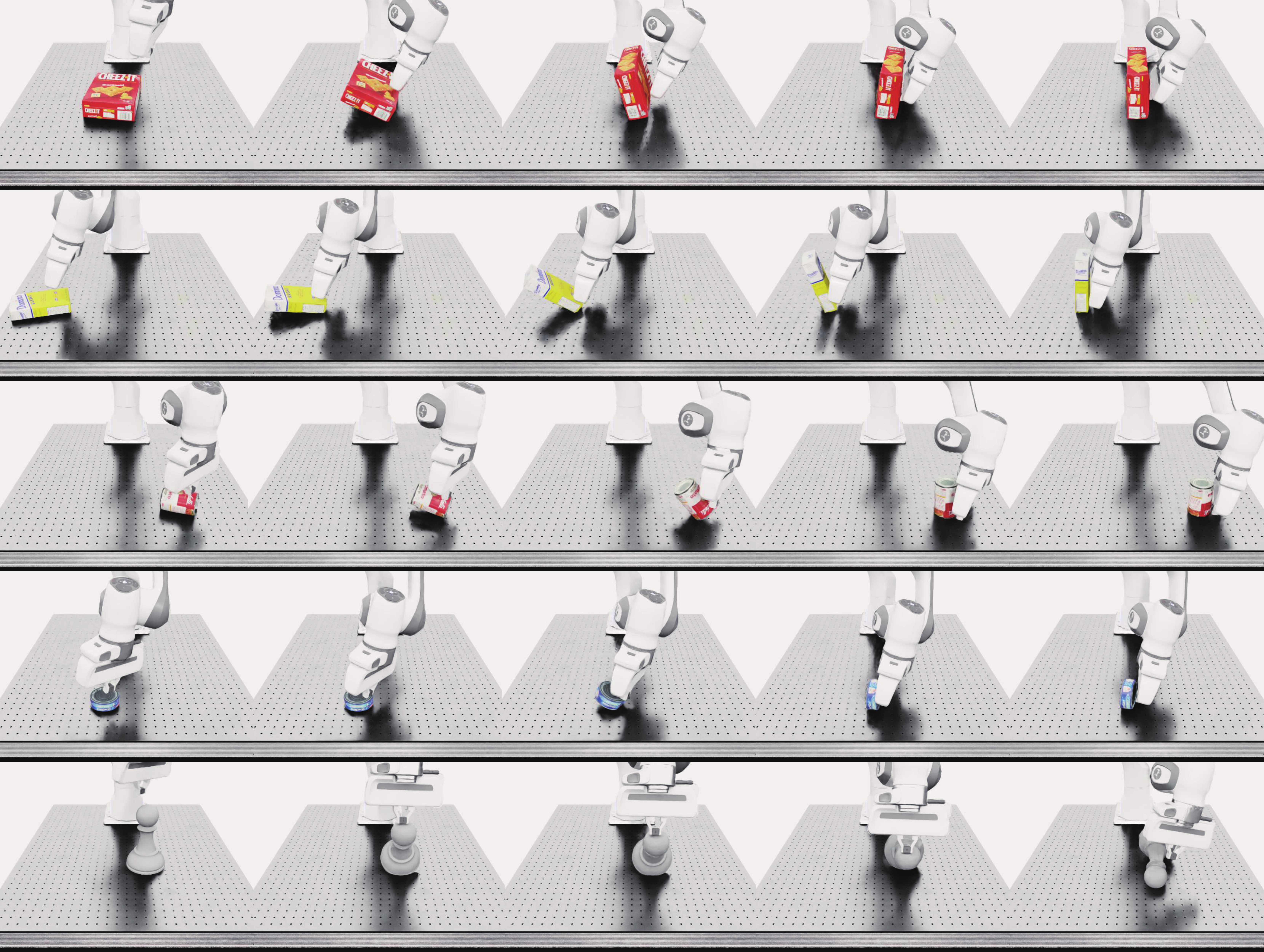}
    \caption{Qualitative results of non-prehensile manipulation in simulation. The policy progressively reconfigures objects through interaction to increase graspability before grasp execution.}
  \label{fig:supp_sim_results}
\end{figure}

We provide additional qualitative examples of the learned manipulation behavior in simulation as shown in~\cref{fig:supp_sim_results}.
Each sequence shows the object being progressively reconfigured through non-prehensile interactions before entering a graspable configuration.

Compared with target pose reaching policies, the learned policy focuses on improving the graspability of the object configuration rather than matching a predefined pose.
As a result, the robot often discovers diverse manipulation strategies such as pushing, sliding, or reorienting the object depending on the initial configuration.

Additional sequences illustrate that the policy can handle a variety of object geometries and initial poses without requiring manually specified target configurations.

\subsection{Additional Qualitative Real World Results}
\label{sec:supp_real_qual}

\begin{figure}[h]
  \centering
    \includegraphics[width=1.0\linewidth]{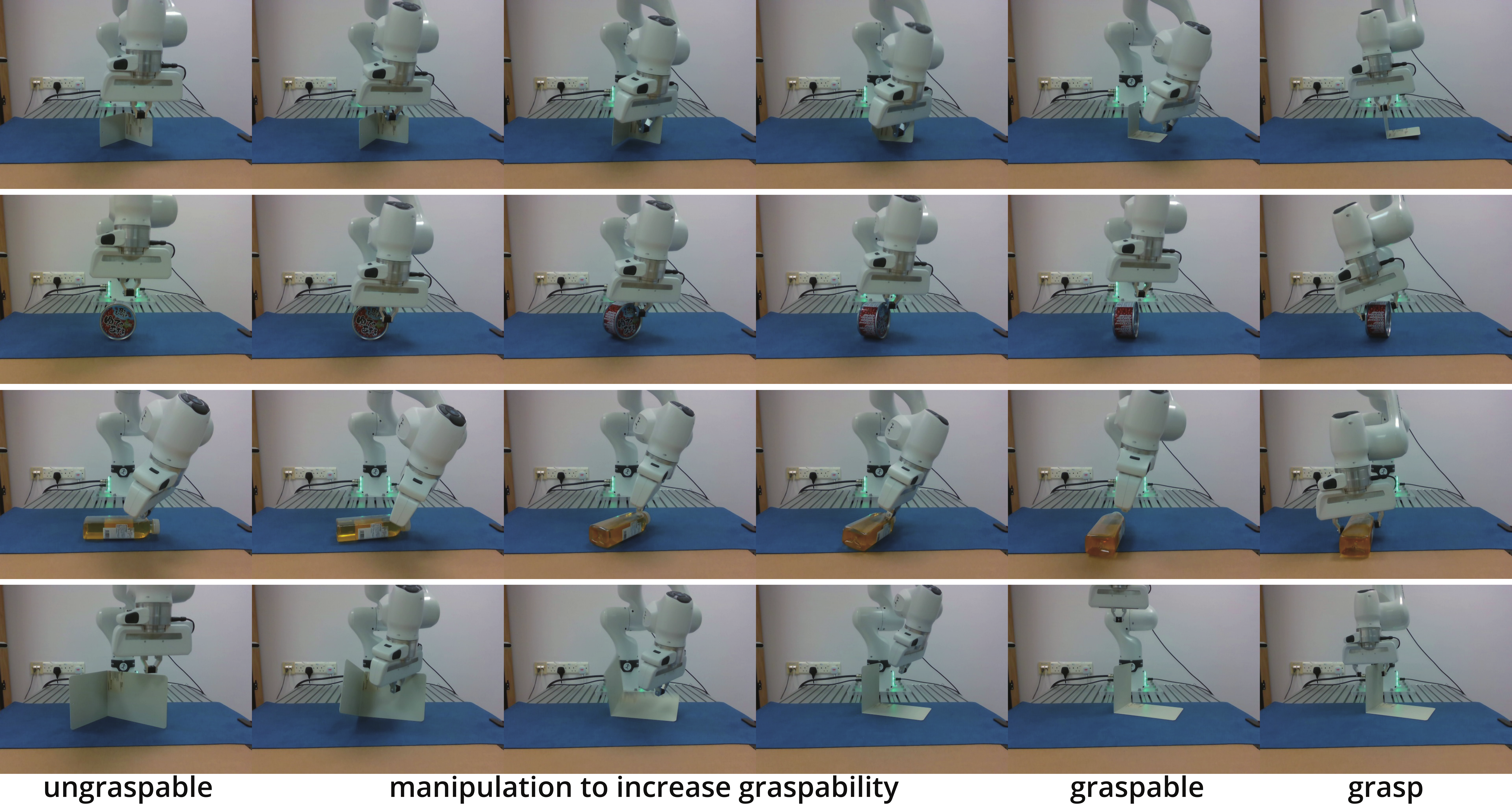}
  \caption{Additional qualitative results of the full manipulation-to-grasp pipeline on real world objects. Each row shows a temporal sequence where the robot manipulates the object to increase graspability and autonomously transitions to grasp execution once a graspable configuration is reached. The graspability is defined related to the home pose of the robot, which is the same for all objects.
  }
  \label{fig:supp_real_results}
\end{figure}

As shown in~\cref{fig:supp_real_results}, we provide additional real world manipulation-to-grasp examples beyond those shown in the main paper.
Each sequence illustrates the full closed-loop pipeline: the robot first performs non-prehensile manipulation to improve graspability and then autonomously transitions to grasp execution.

These examples demonstrate that the learned policy produces consistent manipulation behaviors across objects with different shapes, sizes, and friction properties.
Importantly, the manipulation stage does not rely on explicit pose targets or external planners.
Instead, the transition to grasping is triggered by the predicted graspability distance.

\section{Implementation and Training Details}
\label{sec:supp_impl}

\subsection{Training Pipeline}

The final policy is obtained through a three stage training procedure consisting of teacher reinforcement learning, action range reducing, and student distillation.

\textbf{Stage I: Teacher Policy Training.}
A teacher policy is first trained in simulation using PPO to learn non-prehensile manipulation behaviors.  
This stage uses relatively large action limits to encourage exploration and diverse object interactions.

\textbf{Stage II: Action Range Reducing.}
To stabilize the learned behaviors, we further continue training the teacher policy while gradually tightening the action limits.  
This stage improves control precision and reduces oscillatory manipulation behavior.

\textbf{Stage III: Student Distillation.}
Finally, a recurrent student policy is distilled from the teacher.  
The student operates under partial observations consistent with the real world perception pipeline, while the teacher uses full object observations during training.

The grasp pose head is supervised during student distillation: for each state, we select the nearest graspable reference and transform its precomputed object-frame grasp to the world frame as supervision, with MSE losses on position and 6D rotation.
This head is not used as a manipulation target and does not affect the RL objective.

\subsection{Simulation Environment}

All training is conducted in Isaac Lab using a Franka Panda robot with a parallel-jaw gripper.  
The simulator time step is $\Delta t = 1/80$ s and actions are applied every eight simulation steps, resulting in a control frequency of 10 Hz.  
Each episode lasts 30 seconds.

Object geometry is represented by mesh sampled point clouds. 
For the teacher policy, observations are constructed from the full mesh sampled point cloud. 
For the student policy, observations are obtained from point clouds rendered using nvdiffrast~\cite{laine2020modular}.
A denser set of mesh sampled points is additionally used for graspability related computations.
The simulation setup is summarized in~\cref{tab:supp_sim}.

\begin{table}[t]
\centering
\caption{Simulation parameters used in training.}
\label{tab:supp_sim}
\begin{tabular}{lc}
\toprule
Parameter & Value \\
\midrule
Simulation timestep & $1/80$ s \\
Control frequency & 10 Hz \\
Episode length & 30 s \\
Policy observation points & 512 \\
Grasp-processing points & 1000 \\
\bottomrule
\end{tabular}
\end{table}

\subsection{Domain Randomization}

To improve robustness and facilitate sim-to-real transfer, several physical properties are randomized during training. 
Specifically, we randomize the friction coefficients of the object and the table, the object density, and the object scale. 
These randomizations are applied throughout all training stages, including teacher training, action-range reduction, and student distillation. 
The randomization ranges are listed in~\cref{tab:supp_randomization}.

\begin{table}[t]
\centering
\caption{Domain randomization ranges used in simulation.}
\label{tab:supp_randomization}
\begin{tabular}{lc}
\toprule
Parameter & Range \\
\midrule
Object friction & $[0.3,\,0.7]$ \\
Table friction & $[0.3,\,0.8]$ \\
Object density & $[500,\,1500]$ \\
Object scale multiplier & $[0.8,\,1.2]$ \\
\bottomrule
\end{tabular}
\end{table}

\subsection{Action Space}
The manipulation action is a 6D relative end-effector command executed by a damped-least-squares differential IK controller with the gripper remaining closed.
Actions are clipped to $0.06$\,m translation and $0.1$\,rad rotation, further reduced to $0.02$\,m and $0.03$\,rad during the action-scale curriculum for finer contact control.

\subsection{Reward Formulation}
During teacher policy training, we use a combination of reaching and manipulation rewards to stabilize early exploration and encourage object interaction. The total reward is
\begin{equation}
r_t = w_r r_t^{\text{reach}} + w_m r_t^{\text{manip}},
\end{equation}
where $r_t^{\text{reach}}$ encourages the end-effector to approach the object and 
$r_t^{\text{manip}}$ is derived from the graspability field defined in Sec.~\ref{sec:grasp_field}.
Key reward coefficients are summarized in~\cref{tab:supp_reward}.

\begin{table}[t]
\centering
\caption{Reward formulation coefficients.}
\label{tab:supp_reward}
\begin{tabular}{lc}
\toprule
Coefficient & Value \\
\midrule
$w_r$ & 0.2 \\
$w_m$ & 1.0 \\
$k_1$ & 0.3 \\
$k_2$ & 200.0 \\
$\beta$ & 0.995 \\
$\gamma$ & 0.995 \\
\bottomrule
\end{tabular}
\end{table}

\paragraph{Reaching Reward.}
The reaching component uses potential based reward shaping based on the end-effector distance to the object:
\begin{equation}
r_t^{\text{reach}} =
\gamma \Phi^{\text{reach}}(d_t) -
\Phi^{\text{reach}}(d_{t-1}),
\end{equation}
where $d_t$ denotes the end-effector distance to the object center. 
The potential function follows
\begin{equation}
\Phi^{\text{reach}}(d) = k_1 \beta^{k_2 d}.
\end{equation}

\paragraph{Manipulation Reward.}
The manipulation reward is derived from the graspability distance $\tilde d(s_o)$ defined in Sec.~\ref{sec:grasp_field}. 
Following the potential based shaping used in the main method, the reward encourages reductions in graspability distance between consecutive states:
\begin{equation}
r_t^{\text{manip}} =
\gamma \Phi(\tilde d(s_o^{t+1}))
-
\Phi(\tilde d(s_o^{t})).
\end{equation}

This reward directly encourages the policy to transform the object toward configurations with higher graspability.

\subsection{Curriculum Scheduling}

Two curricula are applied during training to stabilize manipulation behaviors.
The detailed schedules are provided in~\cref{tab:supp_curriculum}.

\begin{table}[t]
\centering
\caption{Curriculum schedules used during teacher training.}
\label{tab:supp_curriculum}
\begin{tabular}{lcc}
\toprule
Curriculum & Parameter & Schedule \\
\midrule
Velocity stability & Linear velocity & $0.3 \rightarrow 0.01$ \\
 & Angular velocity & $1.2 \rightarrow 0.1$ \\
Action scale & Translation limit & $0.06 \rightarrow 0.02$ \\
 & Rotation limit & $0.1 \rightarrow 0.03$ \\
\bottomrule
\end{tabular}
\end{table}

The velocity stability curriculum encourages the object to settle before termination, while the action scale curriculum gradually improves control precision.

\subsection{Policy Architecture}

Both teacher and student policies share the same perception-to-control backbone illustrated in~\cref{fig:pipeline} of the main paper.

Observations are first encoded into object-centric tokens using the encoder introduced in CORN~\cite{cho2024corn}.  
These tokens are processed through cross attention modules conditioned on robot state to obtain a compact representation of the object configuration.

To address partial observability, the student policy incorporates a recurrent module implemented with a gated recurrent unit (GRU)~\cite{cho2014gru}.  
The GRU aggregates temporal information across observations and enables the policy to infer object motion.
The student architecture details are listed in~\cref{tab:supp_arch}.

\begin{table}[t]
\centering
\caption{Student policy architecture.}
\label{tab:supp_arch}
\begin{tabular}{lc}
\toprule
Component & Configuration \\
\midrule
Recurrent module & GRU \\
Number of layers & 2 \\
Hidden dimension & 128 \\
Outputs & action, graspability distance, grasp pose \\
\bottomrule
\end{tabular}
\end{table}

\subsection{Teacher and Student Observations}

Teacher and student policies differ in their observation modalities.
The observation breakdown is shown in~\cref{tab:supp_obs}.

\begin{table}[t]
\centering
\caption{Observation inputs for teacher and student policies.}
\label{tab:supp_obs}
\begin{tabular}{lcc}
\toprule
Observation Component & Teacher & Student \\
\midrule
Robot joint states & \ding{51} & \ding{51} \\
End-effector state & \ding{51} & \ding{51} \\
Previous action & \ding{51} & \ding{51} \\
Object state & \ding{51} & -- \\
Object point cloud & \ding{51} & -- \\
Multi-view partial clouds & -- & \ding{51} \\
\bottomrule
\end{tabular}
\end{table}

The student observations mimic the real world perception pipeline, where partial point clouds are reconstructed from multiple camera views. 
For proprioception, the teacher policy receives robot joint positions and velocities, while the student policy only observes joint positions. 
This design better reflects real world deployment, where control frequencies and sensing intervals may vary, making joint velocity estimates difficult to align with the simulation setting. Temporal information can instead be inferred by the recurrent policy.

\subsection{Optimization}

\begin{table}[t]
\centering
\caption{Key PPO hyperparameters used during teacher training.}
\label{tab:supp_ppo}
\begin{tabular}{lc}
\toprule
Hyperparameter & Value \\
\midrule
Learning rate & $3 \times 10^{-4}$ \\
Discount factor $\gamma$ & 0.99 \\
GAE parameter $\lambda$ & 0.95 \\
PPO clipping parameter & 0.3 \\
Value loss coefficient & 4.0 \\
Entropy coefficient & 0.0 \\
PPO epochs & 8 \\
Rollout length & 10 steps \\
\bottomrule
\end{tabular}
\end{table}

Key PPO settings for teacher model training are summarized in~\cref{tab:supp_ppo}. 
While student distillation optimizes a behavior cloning term and three grasp related prediction terms.
The total loss is
\begin{equation}
\begin{aligned}
\mathcal{L}_{\text{student}}
&=
\mathcal{L}_{\text{behavior}} \\
&+ \lambda_{\text{grasp-prob}} \mathcal{L}_{\text{grasp-prob}} \\
&+ \lambda_{\text{grasp-posi}} \mathcal{L}_{\text{grasp-posi}} \\
&+ \lambda_{\text{grasp-rot}} \mathcal{L}_{\text{grasp-rot}} .
\end{aligned}
\end{equation}

The behavior term matches student actions to teacher actions:
\begin{equation}
\mathcal{L}_{\text{behavior}}
=
\ell_{\mathrm{MSE}}(a_t, a_t^{\text{teacher}}),
\end{equation}
where $\ell_{\mathrm{MSE}}$ denotes mean squared error.

\paragraph{Grasp related predictions.}
In addition to actions, the student policy predicts a grasp pose
\(
\hat{\mathbf{g}}_t = (\hat{\mathbf{p}}_t, \hat{\mathbf{r}}_t)
\),
where $\hat{\mathbf{p}}_t \in \mathbb{R}^3$ is the grasp position and 
$\hat{\mathbf{r}}_t$ is the grasp orientation. 
The orientation is represented using the 6D rotation representation.

The probability distance head uses a Huber loss:
\begin{equation}
\mathcal{L}_{\text{grasp-prob}}
=
\ell_{\text{Huber}}
\left(
\hat d_t,
\tilde d(s_o^t)
\right).
\end{equation}

The grasp position loss is defined as
\begin{equation}
\mathcal{L}_{\text{grasp-pos}}
=
\ell_{\mathrm{MSE}}
\left(
\hat{\mathbf{p}}_t,
\mathbf{p}_t^{*}
\right).
\end{equation}

And the grasp rotation loss is
\begin{equation}
\mathcal{L}_{\text{grasp-rot}}
=
\ell_{\mathrm{MSE}}
\left(
\hat{\mathbf{r}}_t^{6D},
\mathbf{r}_t^{6D,*}
\right).
\end{equation}

Here $\mathbf{p}_t^{*}$ and $\mathbf{r}_t^{6D,*}$ denote the grasp pose targets derived from the reference configuration in the graspable set that is closest to the current object state.
The coefficients used in training are summarized in~\cref{tab:supp_student_loss}.

\begin{table}[t]
\centering
\caption{Student distillation loss settings.}
\label{tab:supp_student_loss}
\begin{tabular}{lc}
\toprule
Setting & Value \\
\midrule
Behavior / position / rotation loss & MSE \\
Grasp-distance loss & Huber \\
$\lambda_{\text{grasp-prob}}$ & 100.0 \\
$\lambda_{\text{grasp-posi}}$ & 100.0 \\
$\lambda_{\text{grasp-rot}}$ & 10.0 \\
\bottomrule
\end{tabular}
\end{table}

\section{Limitations}
Current performance
degrades for highly deformable or extremely low-friction objects that cannot be consistently regulated through contact interactions, and the graspable set depends on coverage of the grasp synthesis model. 
Future work should study learning graspability directly from interaction and extend the framework to cluttered and dynamic environments.

\end{document}